\newcommand{\CLASSINPUTtoptextmargin}{20.1mm}
\newcommand{\CLASSINPUTbottomtextmargin}{19.1mm}
\newcommand{\CLASSINPUTinnersidemargin}{19.1mm}
\newcommand{\CLASSINPUToutersidemargin}{19.1mm}

\documentclass[letterpaper, 10 pt , conference]{IEEEtran}

\IEEEoverridecommandlockouts

\usepackage{graphicx}
\usepackage{mwe}
\usepackage{epsfig}
\usepackage{epstopdf}
\usepackage{mathptmx}
\usepackage{times}
\usepackage{amsmath,amssymb,amsfonts,amsthm}
\usepackage{bm}
\usepackage{mathtools}
\usepackage[dvipsnames]{xcolor}
\usepackage{cite}
\usepackage{algorithmic}
\usepackage{textcomp}
\usepackage[caption=false, font=footnotesize]{subfig}
\usepackage{placeins}

\graphicspath{{figures/}}

\newcommand{\vv}[1]{\mathrm{#1}}

\DeclareMathAlphabet{\mathcal}{OMS}{cmsy}{m}{n}
\newcommand{\transpose}{^\mathsf{T}}

\title{\vspace{18pt} \LARGE \bf Dynamic Modeling of Bucket-Soil Interactions Using Koopman-DFL Lifting Linearization for Model Predictive Contouring Control of Autonomous Excavators*}

\LARGE \bf

\author{Filippos~E.~Sotiropoulos and~H.~Harry~Asada,~\IEEEmembership{Member,~IEEE}%
\thanks{*This material is based upon work supported by the National Science Foundation under Grant No. NSF-CMMI 2021625}
\thanks{F. E. Sotiropoulos (fes@mit.edu) and H. H. Asada (asada@mit.edu) are with the Department of Mechanical Engineering, Massachusetts Institute of Technology, 77 Massachusetts Ave., Cambridge, USA. (The corresponding author is F. E. Sotiropoulos).}
}%

\begin{document}
\maketitle
\begin{abstract}

A lifting-linearization method based on the Koopman operator and Dual Faceted Linearization is applied to the control of a robotic excavator. In excavation, a bucket interacts with the surrounding soil in a highly nonlinear and complex manner. Here, we propose to represent the nonlinear bucket-soil dynamics with a set of linear state equations in a higher-dimensional space. The space of independent state variables is augmented by adding variables associated with nonlinear elements involved in the bucket-soil dynamics. These include nonlinear resistive forces and moment acting on the bucket from the soil, and the effective inertia of the bucket that varies as the soil is captured into the bucket. Variables associated with these nonlinear resistive and inertia elements are treated as additional state variables, and their time evolution is represented as another set of linear differential equations. The lifted linear dynamic model is then applied to Model Predictive Contouring Control, where a cost functional is minimized as a convex optimization problem thanks to the linear dynamics in the lifted space. The lifted linear model is tuned based on a data-driven method by using a soil dynamics simulator. Simulation experiments verify the effectiveness of the proposed lifting linearization compared to its counterpart. 

\end{abstract}

%
%
%
%

\section{Introduction}\label{sec:introduction}

In many earth-moving tasks such as trenching, foundation digging and bench forming, it is necessary to use an excavator to move soil so as to achieve a desired final shape of the site soil. There have been a multitude of attempts from both industry and academia to develop autonomous soil forming robots. Simple autonomous systems have already been commercially available for final grading, in which the very final layer of soil is removed to create a precise soil shape. This yields a precise final soil shape but does not deal with the most challenging issue of precision digging: at deeper digging depths, the excavator experiences large, nonlinear forces that arise from the dynamic bucket-soil interactions. This is often cited as one of the most prominent challenges\cite{Dadhich2016} and various approaches have been proposed to tackle this.

Both model-free and model-based approaches have been employed. Singh \cite{Singh1995} used the Fundamental Earthmoving Equation (FEE) and a learned model to constrain the action space from which an optimal trajectory is planned. More recently, Yang et al. \cite{Yang2020TimeVM} have used an optimization approach with a simple analytical soil model to determine time varying minimum torque trajectories.

Limitations in available models have led to efforts to reduce the necessity for a model altogether. One common approach has been to use interaction control to regulate the forces between bucket and soil. An early effort employed by Bernold \cite{Bernold1993} was to use impedance control so as to control the relationship between the bucket trajectory error and the exerted force. In this vein, Richardson-Little and Damaren \cite{Little2005} utilized a rheological model in combination with a compliance controller. An alternate approach proposed by Jud et al. \cite{Jud2017} is to use a desired force trajectory to maintain a desirable interaction with the soil.

While model free control methods can react to changing soil forces, they are limited in achieving precise path tracking. When precise tracking of a path is required, it is necessary to predict interaction forces that vary dynamically depending on nonlinear soil properties as well as on the changing soil profile. One must predict and compensate for the complex nonlinear forces. Model Predictive Control (MPC), a framework for realizing prediction and compensation, meets this goal. The challenge, however, is to construct a dynamic model to predict the highly nonlinear and complex bucket-soil interaction. 

For the use of MPC, we seek a bucket-soil interaction model that meets several key requirements. Specifically, the model must:
\begin{itemize}
    \item Capture the rich nonlinear bucket-soil dynamics;
    \item Be trainable from sampled trajectory data;
    \item Exploit sensing, including visual mapping;
    \item Be usable for real-time optimization; and
    \item Account for the effect of soil shape.
\end{itemize}

To fulfil these requirements, we aim to exploit an emergent modeling methodology, termed lifting linearization. This modeling methodology, underpinned by the theory of the Koopman operator \cite{Koopman_1931_original} and grounded in physical modeling theory and Dual Faceted Linearization (DFL) \cite{dfl1_dsmc}, can capture complex nonlinear dynamics with a set of linear differential equations in a higher-dimensional space. The linear representation of the nonlinear dynamics can remove or alleviate the fundamental difficulty in the use of MPC for real-time control. The cost functional of MPC can be minimized as a convex optimization problem subject to linear dynamics and linear constraints. 

In the following, Koopman operator and DFL will be briefly summarized for readability, and the nonlinear dynamics of bucket-soil interaction will be lifted and represented with higher-dimensional linear equations. MPC will then be applied to the lifted linear system. The path tracking problem will be formulated as a contouring control form of MPC \cite{Lam2010}, where the bucket can be controlled along a task defined spatial path while satisfying constraints relating to the task and actuation limits. Simulation experiments evaluate the modeling accuracy and demonstrate the contouring control performance.

\section{Lifting Linearization}\label{sec:dfl}

Lifting linearization is rooted in the seminal work by Koopman \cite{Koopman_1931_original}. His operator theory underpins more recent development of lifting linearization. The Koopman Operator, $\mathcal{K}$ is a linear infinite dimensional operator which evolves observations $g(\vv{x})$ of the state $\vv{x}\in \mathbb{R}^n$ of an autonomous nonlinear dynamical system such that:
\begin{equation}
    \frac{d}{dt}g(\vv{x}) = \mathcal{K}g(\vv{x}) 
\end{equation}
or
\begin{equation}
    g(\vv{x}_{k+1}) = \mathcal{K}_\mathrm{d} g(\vv{x}_k) 
\end{equation}
for the continuous- and discrete-time systems with no control, respectively.

\subsection{Approximating the Koopman Operator for systems with Control}
Being infinite-dimensional and for purely autonomous systems means that the Koopman operator has limited utility in the context of practical nonlinear control systems. To remedy this, a finite-dimensional approximation of the Koopman operator may be obtained as well as an approximation of the closest linear effect of the exogenous inputs such that:
\begin{equation}
    \xi_{k+1} = \mathrm{A} \xi_{k} + \mathrm{B} \vv{u}_k
\end{equation}
where $\vv{u}_k\in\mathbb{R}^{n_u}$ is the input, and $\xi_k = [\vv{x}_k\transpose \; g_1(\vv{x}_k) \; \dots \;  g_{n_g}(\vv{x}_k)]\transpose\in\mathbb{R}^{n+n_g}$ is the set of chosen observation functions including the state itself for convenience. The discrete transition matrices with compatible dimensions $\mathrm{A}$ and $\mathrm{B}$ can then be regressed directly from $N_D$ data:
\begin{equation}\label{eq:koop_regression}
   \mathrm{G}^* = \mathrm{\Xi}\mathrm{\Upsilon}^\dagger
\end{equation}
where $\mathrm{G}^* = [\mathrm{A} \; \mathrm{B}]$,   $\mathrm{\Xi} = [\xi_1 \; \xi_2  \; \dots  \; \xi_{N_D}]$ is the measured state and observables data matrix, $\mathrm{\Upsilon} = [\vv{y}_0 \; \vv{y}_1\; \dotsb  \; \vv{y}_{N_D -1}]$ is the regressor data matrix with $\mathrm{y}_k= [\xi_k\transpose\; \vv{u}_k\transpose]\transpose $ and $\mathrm{\Upsilon}^\dagger$ is the Moore-Penrose pseudo-inverse of $\mathrm{\Upsilon}$.

This outline is only the briefest introduction to the concept of identifying linear representations of nonlinear control systems based on the Koopman Operator, but for a complete treatment the reader is referred to \cite{MezicAppliedKoopmanism2012} as well as several examples of this idea being applied to robotic systems \cite{Abraham2017}.

\subsection{Observables for Lifting Linearization}

One factor not mentioned thus far is what functions are to be used as the observables. This is a critical factor in applying lifting linearization, as a good selection of observables can not only improve the modeling accuracy but also do so with a lower order model. To this end, different approaches have been proposed, including identifying the best observables from large libraries of nonlinear observables\cite{Kaiser_2021}, using generic observables such as radial basis functions \cite{Korda_2018_MPC_koopman}, harnessing the kernel trick to efficiently represent the nonlinear observables \cite{Williams_2015_Koopman_kernel}, using an optimization framework to determine Koopman eigenfunctions \cite{Korda_2020_Optimization_koopman}, and even using a learning approach to learn the nonlinear dictionary terms \cite{Li_2017_koopman_learning}.
A point that has been noted by multiple authors is that choosing observables that are either related to the form of the nonlinearity in the differential equation if known \cite{ Wilcox_2019_Bilinear} or observables that are otherwise physically motivated \cite{Korda_2018_MPC_koopman} often leads to superior results. 

In this vein, Dual-Faceted Linearization (DFL) \cite{dfl1_dsmc, dfl2_acc} is a  method of lifting linearization which exploits principles from physical modeling theory, in particular, the structure of a lumped parameter model, to determine the observables. DFL determines a special set of measurements used as lifting variables (termed ``auxiliary variables'') in a systematic manner based on the connectivity of physical elements depicted in a bond graph \cite{Karnopp2006} or by inspecting the equations of motion of the system. By construction, the auxiliary variables all relate to physical quantities in the original nonlinear system and thus can often be \textit{directly measured}.  Thus, even without knowledge of the system's nonlinear constitutive relations one can create the structure of the lifted linear system, and allow for regressing the linear model using the physical measurements of the states \textit{and} the specified auxiliary variables. 

\subsection{Lifting Linearization based on DFL}
In this section, we outline the essentials to generating DFL models. An in-depth treatment on this subject can be found in \cite{dfl1_dsmc} but an overview is presented here for clarity.

The essence of the DFL method is defining the special lifting variables, termed auxiliary variables  $\eta \in \mathbb{R}^{n_a}$, as the \textit{output} of \textit{all} nonlinear elements involved in the nonlinear dynamical system. 
This representation is motivated by the natural linearity arising from the connectivity of elements in physical systems. Connections of elements, be they linear or nonlinear, are governed by linear relations. For example, Kirchoff's current law says that all currents going in and out of a point sum to zero. Similarly, in the mechanical domain, Newton's second law dictates that all the forces acting on a body (including the inertial force) sum to zero. 

Using this definition results in the system being modeled by the following set of linear differential equations:
\begin{align}\label{eq:dfl} 
\begin{bmatrix}
        \dot{\vv{x}}\\ \dot{\vv{\eta}}
    \end{bmatrix} = \underbrace{\begin{bmatrix}
         \mathrm{A}_{c,\vv{x}} &  \mathrm{A}_{c,\vv{\eta}} \\
         \mathrm{H}_{c,\vv{x}} &   \mathrm{H}_{c,\vv{\eta}}
    \end{bmatrix}}_{\mathrm{A}_c} \begin{bmatrix}
        \vv{x}\\ \vv{\eta}
    \end{bmatrix} +  \underbrace{\begin{bmatrix}
        \mathrm{B}_{c,\vv{x}} \\ \mathrm{H}_{c,\vv{u}}
    \end{bmatrix}}_{\mathrm{B}_c}\vv{u}
\end{align} where $\mathrm{A}_{c,\vv{x}}\in \mathbb{R}^{n \times n}$, $\mathrm{A}_{c,\vv{\eta}}\in \mathbb{R}^{n \times n_a}$ and $\mathrm{B}_{c,\vv{x}}\in \mathbb{R}^{n \times n_u}$ are exact parameter matrices determined by the system structure, whereas $\mathrm{H}_{c,\vv{x}} \in \mathbb{R}^{n_a \times n}$, $ \mathrm{H}_{c,\vv{\eta}} \in \mathbb{R}^{n_a \times n_a}$, and $ \mathrm{H}_{c,\vv{u}}\in \mathbb{R}^{n_a \times n_u}$ are approximate and to be determined from data.

The exactness of $\mathrm{A}_{c,\vv{x}}$, $\mathrm{A}_{c,\vv{\eta}}$ and $\mathrm{B}_{c,\vv{x}}$, which is proven in \cite{dfl1_dsmc}, is facilitated by the definition of the auxiliary variables which are defined as the outputs of all the nonlinear elements in a lumped-parameter system. A similar idea is presented in \cite{Folkestad2020} for Lagrangian systems where the position variables have an exact differential equation. 

The estimated matrices $\mathrm{H}_{c,\vv{x}}$, $ \mathrm{H}_{c,\vv{\eta}}$, and $ \mathrm{H}_{c,\vv{u}}$ can then be regressed from data using a least squares estimate:
\begin{equation}\mathrm{H}= \arg \mathop {\min }\limits_\mathrm{H} \mathbb{E}\left[ {{{\left\| {\dot \eta - {{\dot \eta }_{m}}} \right\|}^2}} \right]\end{equation}
where $\dot{\eta}_m$ are measured data of the auxiliary variable time derivatives and $\mathrm{H} = [\mathrm{H}_{c,\vv{x}} \;  \mathrm{H}_{c,\vv{\eta}} \;  \mathrm{H}_{c,\vv{u}}]$.

This methodology has been compared with more typical Koopman operator modeling and it was reported that choosing the observables based on DFL yielded performance comparable to a Koopman model utilizing a substantially larger vector of observables \cite{Igarashi2020}. This makes DFL advantageous in applications where having a compact and interpretable model is necessary.

In a control setting we may be interested in utilizing a discrete representation of the system. This can be derived through using an ordinary system discretization: $A = e^{A_c \Delta t}$  and $B = A_c^{-1}(A-I)B_c$  from the continuous DFL model or directly regressing the system matrices exactly as in (\ref{eq:koop_regression}) where $\xi_k=[\vv{x}_k\transpose \eta_k\transpose]\transpose$.

In this work, we will use the DFL method for selecting observable variables and constructing the lifting linearization. This is because it offers some specific features and benefits:
\begin{itemize}
    \item The model is composed of physically meaningful variables. As such, the model is accountable and provides the control designer with physical intuitions. 
    \item The lifting variables can be measured with sensors, providing richer data than independent state variables alone. It opens up an avenue for exploiting sensor technology for improved modeling and control. 
    \item Choosing the observables is structured, reducing trial-and-error efforts for selecting lifting variables. 
\end{itemize}

\section{Excavation Model with DFL}
In this section, we present the modeling of the bucket-soil system using DFL. In the spirit of simplicity and to elucidate the capability of the lifted linearization to capture the soil dynamics, we purely model and control a bucket while decoupling the boom-arm structure and dynamics. This equates to controlling a robot in operational space \cite{Khatib1987}. Although challenging due to high levels of internal friction in excavation machinery, there have been several successful endeavours in controlling excavators in task space \cite{Maeda2011,Maeda2015,Jud2017}.

In addition, we consider the bucket moving purely in the vertical plane. This is a practical simplification since the forces necessary to move a bucket sideways generally mean these motions are either not possible or practical.

\subsection{Bucket State Equations}

Given the high level assumptions, the bucket is modeled to translate and rotate in the vertical plane (see Fig \ref{fig:system_states}). As the soil ahead of the bucket fails through shearing, it exerts the forces $e_{T,x}$ and  $e_{T,z}$ on the bucket. These nonlinear forces also generate a moment $e_{T,\phi}$ on the bucket. Physically, these forces and moment are generated through highly nonlinear bucket-soil interactions, which are nonlinear resistive elements. According to the DFL method, $e_{T,x}$, $e_{T,z}$, $e_{T,\phi}$ are auxiliary variables, since they are the outputs of nonlinear resistive elements.

These forces and moments in conjunction with the control inputs, which are the actuator forces and torque, $\vv{u} = [u_x \; u_z \; u_\phi]\transpose$ , represent the total forces and torque acting on the bucket. From Newton's Second Law, the total forces and moment are equal to the time rate of change to the momenta and angular momentum of the bucket and the soil that moves together with the bucket.

\begin{equation}\label{eq:nonlinear_system_bucket_momenta}
\begin{aligned}
    \dot{p}_x       & = -e_{T,x} + {u_x} \\
    \dot{p}_z       & = -e_{T,z} + {u_z} \\
    \dot{p}_\phi    & = -e_{T,\phi} + {u_\phi}
\end{aligned}
\end{equation}
where $p_x$, $p_z$, $p_\phi$ are, respectively, momenta and angular momentum of the combined bucket and soil that move together. According to the physical modeling theory and the DFL modeling, the output of an inertial element is velocity. As the inertia of the soil captured by the bucket is not constant, the functional relationship between momenta and velocities of the combined bucket and soil is nonlinear. This implies that the velocities are auxiliary variables, according to the DFL formalism.
Let $x$ and $z$ be the coordinates of the tip of the bucket and $\phi$ the orientation of the bucket, as shown in Fig. \ref{fig:system_states}. Their time derivatives are:
\begin{equation}\label{eq:nonlinear_system_bucket_velocities}
\begin{aligned}
    \dot{x}         & = v_x \\
    \dot{z}         & = v_z \\
    \dot{\phi}      & = \omega \\
\end{aligned}
\end{equation}
Equations \eqref{eq:nonlinear_system_bucket_momenta} and \eqref{eq:nonlinear_system_bucket_velocities} represent the state equations of the bucket dynamics, where the independent state variables are $\vv{x} = [x \; z \; \phi \; p_x \; p_z \; p_\phi ]\transpose$. Note that the state equations are linear equations of the auxiliary variables $v_x$, $v_z$, $\omega$ and $e_{T,x}$, $e_{T,z}$, $e_{T,\phi}$ as well as inputs. These correspond to the upper half of the lifted state equations \eqref{eq:dfl}.

\subsection{Dynamics of auxiliary variables}
\begin{figure}[hbt!]
    \centering
    \includegraphics[trim={1.5cm 6.8cm 14.3cm 8cm},clip, width=0.89\columnwidth]{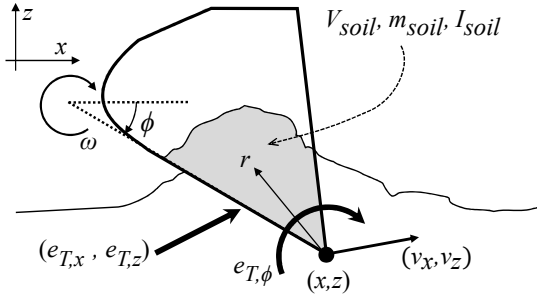}
    \caption{Variables used as states and auxiliary variables in the planar excavation model.}\label{fig:system_states}
\end{figure}

In lifting linearization, nonlinear elements, such as the resistive element associated to the bucket-soil interactions and the interial elements associated to the combined bucket-soil system, are linearized not merely by taking an algebraic approximation but by recasting the nonlinear dynamics in a higher dimensional space and approximating it as a higher-order linear differential equation. Auxiliary variables $e_{T,x}$, $e_{T,z}$, $e_{T,\phi}$ for example, are treated as state variables possessing linear state equations, as in the lower half of equation \eqref{eq:dfl}. 
As for the auxiliary variables, $v_x$, $v_z$, $\omega$, associated to the nonlinear inertia of the combined bucket and soil, the nonlinearity comes from the varying mass and moment of inertia.
\begin{equation}\label{eq:momenta}\begin{aligned}
  v_x       & = p_x / m,    &   v_z  & = p_z / m \quad &\textrm{where }  m = m_{bucket}+m_{soil} \\
  \omega    & = p_\phi / I, &        &                  &\textrm{where }  I = I_{bucket}+I_{soil}.
\end{aligned}\end{equation}
Note that the bucket inertias, $m_{bucket}$ and $I_{bucket}$, are constant and that soil inertia, $m_{soil}$ and $I_{soil}$ are variables that dynamically vary as the soil is collected into the bucket. We treat these variables as auxiliary variables and represent their dynamics as a linear differential equation of the motion of the bucket interacting with soil. As shown in Fig. \ref{fig:system_states}, the transition of $m_{soil}$ and $I_{soil}$ depends on a) bucket position, b) bucket velocity, c) soil surface profile, d) the amount and distribution of soil captured by the bucket and e) soil properties. Variables a) are involved in the bucket state $\vv{x}$, b)  is in the auxiliary variables $\eta$, d) is represented with the current soil mass and moment of inertia $m_{soil}$ and $I_{soil}$ which are now part of the auxiliary variables.
\begin{equation}
    \eta =  [v_x  \; v_z  \; \omega \;  e_{T,x} \;  e_{T,z} \; e_{T,\phi}\;  m_{soil} \; I_{soil} ]\transpose \in \mathbb{R}^8
\end{equation}
Parameters of soil properties are assumed constant. The remaining variables are related to the soil surface profile c).

\subsection{Inclusion of Soil Shape}

\begin{figure}[hbt!]
    \centering
    \includegraphics[trim={3.3cm 7.2cm 5.8cm 5.1cm},clip, width=0.95
    \columnwidth]{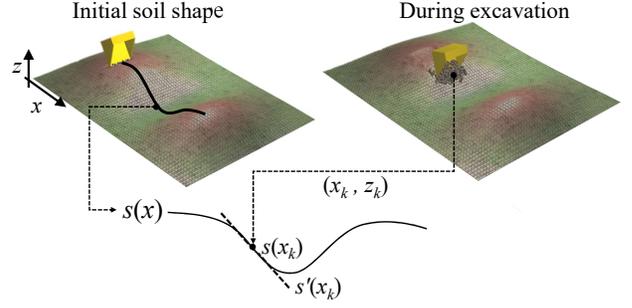}
    \caption{The soil shape $s(x)$ is illustrated on AGX Dynamics simulation environment. $\vv{s}_k$ is determined by evaluating what the soil height and gradient of the initial soil shape was at the current bucket tip location $x_k$.}\label{fig:soil_shape}
\end{figure}

We represent the soil profile as a function of $x: s=s(x)$. See Fig. \ref{fig:soil_shape}. The soil profile $s(x)$ influences the bucket dynamics. Variation in the profile results in changes in the resistive forces and moments, as well as in the soil captured by the bucket. Therefore, we treat the soil profile as an exogenous input or disturbance, and assume that the soil profile input drives the system linearly. 
\begin{equation}
    \xi_{k+1} = \mathrm{A} \xi_{k} + \mathrm{B}\vv{u}_k +\mathrm{B}_s \vv{s}_k 
\end{equation}
where $\vv{s}_k=[s(x_k)\; s'(x_k) \;]\transpose \in \mathbb{R}^2$. Numerical analysis reveals that the soil height $s$ and its gradient $s'$ have significant influences upon the system dynamics, while the influence of its higher-order spatial derivatives are negligible.  The auxiliary variable dynamics can be determined in the same ways as (\ref{eq:koop_regression}) where now  $\mathrm{G}^* \triangleq [\mathrm{A} \; \mathrm{B} \; \mathrm{B}_s]$ and $\mathrm{y}_k = [\vv{\xi}_k\transpose \vv{u}_k\transpose\; \vv{s}_k\transpose]\transpose $. 

\section{Control}\label{sec:control}

Based on the lifted dynamic model, this section aims to construct a controller for tracking a geometric path.

\subsection{Model Predictive Contouring Control}
In an excavation task, the high level objective is to have the bucket pass through a certain path relative to the surface of the soil so as to shape the soil in a specific manner. In this sense, the control objective is not to have the bucket follow a time trajectory of positions but rather to traverse the geometric and spatial path as closely as possible. Thus, the predictive control methodology used to determine the input  $\vv{u}$, is based on the model predictive contouring control (MPCC) algorithm presented by Lam et al. \cite{Lam2010}.
\begin{figure}[hbt!]
    \centering
    \includegraphics[trim={2.4cm 5.3cm 12cm 11cm},clip, width=0.99\columnwidth]{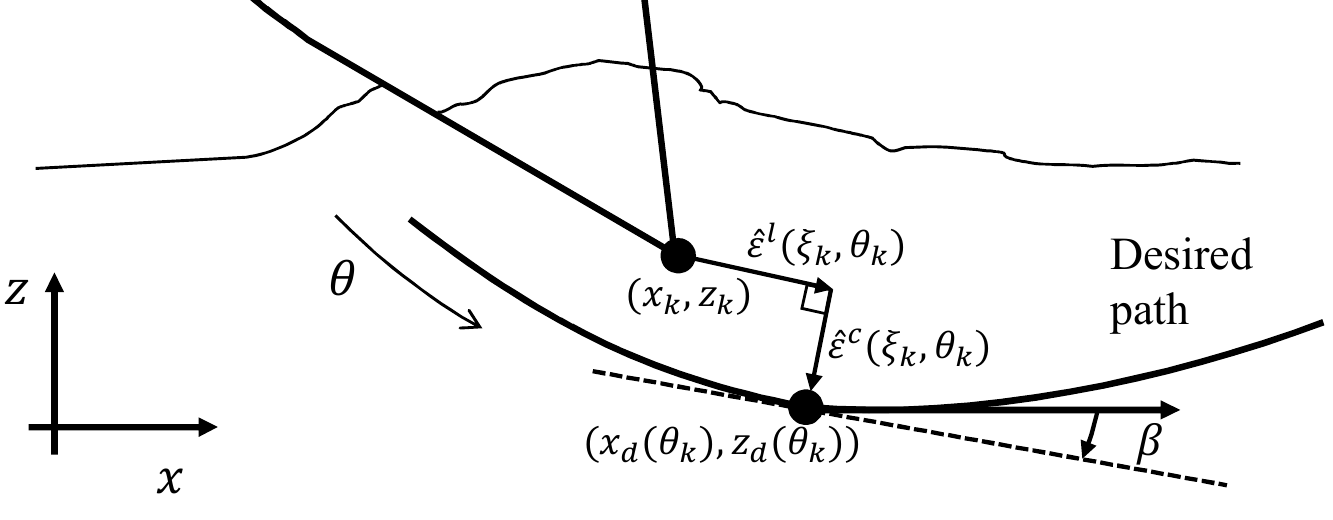}
    \caption{The cost of the MPCC is composed of two error, $\hat{\epsilon}^c_{k}$ the contouring error and $\hat{\epsilon}^l_{k}$ the lag error between the bucket tip and a point on the path parametrized by $\theta_k$.}\label{fig:mpcc_error_diagram}
\end{figure} 

The crux of the methodology is to augment the system dynamics with a virtual state $\theta$, which represents the arc-length of a given path. As shown in Fig. \ref{fig:mpcc_error_diagram}, virtual state $\theta$  dictates the desired position on the path $(x_d(\theta),z_d(\theta))$ for each trajectory timestep. Instead of calculating the shortest distance between path and bucket tip, which is computationally expensive, the path error is represented based on two orthogonal components, the contouring and lag errors, as shown in Fig. \ref{fig:mpcc_error_diagram}.
\begin{align}
&\begin{aligned}\label{eq:contouring error}
\hat{\epsilon}^c_{k}(\vv{\xi}_{k},\theta_{k})=  & \phantom{-}\sin \beta(\theta_k)(x_k-x_d(\theta_k))\\  &-\cos \beta(\theta_k)(z_k-z_d(\theta_k))\end{aligned}\\
&\begin{aligned}\label{eq:lag error}
\hat{\epsilon}^l_{k}(\vv{\xi}_{k},\theta_{k}) = &-\cos \beta(\theta_k)(x_k-x_d(\theta_k)) \\  &-\sin \beta(\theta_k)(z_k-z_d(\theta_k))\end{aligned} 
\end{align}
where, 
\begin{equation}
    \beta(\theta_k) = \arctan \left(\frac{\partial z_d /\partial \theta }{\partial x_d / \partial \theta } \bigg|_{\theta=\theta_k} \right)
\end{equation}
and the dynamics of the path variable $\theta$ are controlled by the virtual input $\upsilon_k$:\begin{equation}
    \theta_{k+1} = \theta_k + \upsilon_k,\quad \upsilon_k\in [0,\upsilon_{max}], \quad \upsilon_{max}> 0
\end{equation}

In  the MPCC formulation, these contouring and lag errors are evaluated over a given time horizon, say $N$ time-steps. While the dynamics of the system have been expressed linearly in the lifted space, the transformation of the contouring and lag errors to the state variables is nonlinear, as shown in eqs. \eqref{eq:contouring error} and \eqref{eq:lag error}. Furthermore, the soil profile $s(x_k)$ is a nonlinear function of $x_k$. 

For both the errors, $\hat{\epsilon}^c_{k}(\vv{\xi}_{k},\theta_{k})$ and $\hat{\epsilon}^l_{k}(\vv{\xi}_{k},\theta_{k})$, and the soil surface shape $\mathrm{s}_k$,  a linearization is performed about the optimized trajectory at the previous time-steps:  $\hat{\mathrm{\Xi}}^*_k = \{ \hat{\vv{\xi}}^*_{k,k},\ \dots ,\ \hat{\vv{\xi}}^*_{k+N-1,k} \}$ , and  $\hat{\Theta}^*_k = \{ \hat{\theta}^*_{k,k},\ \dots ,\ \hat{\theta}^*_{k+N-1,k} \}$. This results in the linearized contouring and lag error:
\begin{equation}
    \begin{aligned}
    \hat{\epsilon}^{a,c}_{k+i,k} &=   \hat{\epsilon}^{c}(\hat{\vv{\xi}}_{k+i,k}^*,\hat{\theta}_{k+i,k}^*) + \nabla \hat{\epsilon}^{c}(\hat{\vv{\xi}}_{k+i,k}^*,\hat{\theta}_{k+i,k}^*) \begin{bmatrix}\vv{\xi}_{k+1} \\ \vv{\theta}_{k+1}
    \end{bmatrix}\\
    \hat{\epsilon}^{a,l}_{k+i,k} &=   \hat{\epsilon}^{l}(\hat{\vv{\xi}}_{k+i,k}^*,\hat{\theta}_{k+i,k}^*) + \nabla \hat{\epsilon}^{l}(\hat{\vv{\xi}}_{k+i,k}^*,\hat{\theta}_{k+i,k}^*) \begin{bmatrix}\vv{\xi}_{k+1} \\ \vv{\theta}_{k+1}
    \end{bmatrix}
    \end{aligned}
\end{equation} and linearized soil profile:
\begin{equation}\begin{aligned}
    \vv{s}_{k+i-1,k}^a &= \begin{bmatrix}
        s( \hat{x}_{k+i-1,k}^*) \\
        s'( \hat{x}_{k+i-1,k}^*)
    \end{bmatrix} +   \begin{bmatrix}
        s'( \hat{x}_{k+i-1,k}^*) \\
        s''( \hat{x}_{k+i-1,k}^*)
    \end{bmatrix} (x_{k+i-1} - \hat{x}_{k+i-1,k}^*)\\  i & = 1,2, \dots, N
\end{aligned}\end{equation}

With these linearized errors and dynamics, the following convex QP can be defined to solve the MPCC problem:
\begin{equation}
\begin{aligned}
\min_{\substack{\vv{u}_{k+i-1}, \; \upsilon_{k+i-1} \\ \vv{\xi}_{k+i},\; \theta_{k+i}\\           i = 1,2, \: \dots \:, N} } \quad & J_k \\
\textrm{s.t.} \quad & 
    \xi_{k+i} = \mathrm{A} \xi_{k+i-1} + \mathrm{B} \vv{u}_{k+i-1} + \mathrm{B}_s\vv{s}_{k+i-1,k}^a\\
                    & \vv{\theta}_{k+i} =  \vv{\theta}_{k+i-1} + \upsilon_{k+i-1} \\
                    & \vv{u}_{k+i-1} \in [\vv{u}_{min},\vv{u}_{max}], \quad \upsilon_{k+i-1}\in[0, \upsilon_{max}]\\
                    &  \vv{\xi}_{k+i} \in [\vv{\xi}_{min},\vv{\xi}_{max}] , \quad \quad \theta_{k+i}\in[\theta_s,0]\\
\end{aligned}
\end{equation}

where the cost $ J_k$ at time $k$, is:
\begin{equation*}\begin{aligned}
    J_k &= \sum^N_{i=1} \bigg( \begin{bmatrix} \hat{\epsilon}^{a,c}_{k+i,k} \\
                               \hat{\epsilon}^{a,l}_{k+i,k}
    \end{bmatrix} \transpose Q \begin{bmatrix} \hat{\epsilon}^{a,c}_{k+i,k} \\
                               \hat{\epsilon}^{a,l}_{k+i,k}
    \end{bmatrix} -q_\theta \theta_{k+1}  \\
     &\phantom{XXXXXXXXXXXX}  + \begin{bmatrix} \Delta \vv{u}_{k+1} \\ \Delta \upsilon_{k+1}
    \end{bmatrix} \transpose R\begin{bmatrix} \Delta \vv{u}_{k+1} \\ \Delta \upsilon_{k+1}
    \end{bmatrix}
    \bigg)
\end{aligned}\end{equation*}
where $Q \in \mathbb{R}^{2\times2}$ defines the contouring and lag cost, $q_\theta\in \mathbb{R}$ rewards progression along the path, $R \in \mathbb{R}^{4\times4}$ defines the cost of changes in the input and $N$ is the prediction horizon.

One of the key benefits of using an optimization framework to control our system is that we can set constraints on the control inputs and states. Here for the majority of the states we set the limits so that the states fall within the 10-th and 90-th percentile of the training data used to determine the model. This is especially important since the linear model of the nonlinear system will be most valid in the region of the state space where it was trained. 

Additionally we can set state constraints which may aid in the excavation task. Firstly, we constrain the horizontal velocity to be positive: $ 0 \leq v_{x,k}$. The vertical position of the bucket tip is constrained to be below the surface of the soil according to $s(x)$ and the previous optimized trajectory such that $ z_{k+i} \leq s( \hat{x}_{k+i,k}^*) $.

Furthermore, DFL allows us to constrain the auxiliary variables. Unlike ordinary Koopman modeling where the observables do not possess any specific physical interpretation, in DFL the auxiliary variables (lifting variables) are physically meaningful and thus the system designer may seek to constrain them. In excavation this may be applied in several places. Firstly, the forces on the bucket can be directly constrained, for example, so as to reduce wear and damage to the bucket. Moreover, since the soil mass, $m_{soil}$,  is part of the lifted space, one can constrain it to within desired amounts so as to collect a certain amount of soil.

\section{Numerical System Simulation}
This section aims to implement and test the proposed method in a simulation environment based on the \textit{AGX dynamics}\footnote{https://www.algoryx.se/agx-dynamics/} physics simulator and the specialized module for the simulation of machine-soil interactions: \textit{agxTerrain} \cite{Servin2021}. This simulation environment allows for testing our proposed method on a system which embodies many of the characteristic nonlinearities present in an excavation system. The MPCC QP is solved using the OSQP solver\footnote{https://osqp.org/} \cite{osqp}.

\subsection{Digging environment}
The soil shape is randomized by adding and subtracting multiple 2-D Gaussians (with randomly sampled height and standard deviation) from a height field with a randomized gradient. This yields soil surfaces with greatly varying shapes (See Fig. \ref{fig:soil_shape} for a sample site shape). The soil properties are implemented using the preset calibrated ``gravel" material model. 

\subsection{Data Collection}
In most examples in the lifting linearization literature, open loop random signals are applied to the system inputs to collect training data. However, this is not feasible in this application. Controlling the bucket without feedback readily results in either becoming deeply lodged in the soil and stalling, or leaving the soil altogether. Instead, we implement a simple PID control scheme whereby the translational d.o.f are speed controlled, and the angle of the bucket is position controlled: 
\begin{equation}
    \begin{aligned}
    u_x & = PID(v_x - \mathcal{U}(v_{x,min},v_{x,max})) + \mathcal{U}(-w_{x},w_{x}) \\ 
    u_z & = PID(v_z - \mathcal{U}(v_{z,min},v_{z,max})) + \mathcal{U}(-w_{z},w_{z}) \\ 
    u_\phi & = PID(\phi - \mathcal{U}(\phi_{min},\phi_{max})) + \mathcal{U}(-w_{\phi},w_{\phi}) \\ 
    \end{aligned}
\end{equation} The setpoints are drawn from a uniform distribution ($\mathcal{U}(a,b)$ between $a$ and $b$)  and additionally random noise, again from a uniform distribution, is added. This injection of noise is necessary so as to be able to correctly identify the system dynamics despite utilizing feedback to collect the data \cite{Proctor2018}. The control and data sampling frequency is 30Hz.

\subsection{Measurement of lifted state variables}

Modeling and control based on DFL lifting linearization requires the access to all the states and auxiliary variables. This section discusses how these variables can be obtained in both simulation environment and experimental setting. The bucket positions $(x, \; z, \; \phi)$ and velocities $(v_x,  \; v_z,  \; \omega)$ are accessible in both simulation and experiment. 
The interactive bucket-soil forces $e_{T,x}$, $e_{T,z}$ and $e_{T,\phi}$ can be extracted from the internal  variables in the simulation environment described above. The inertial variables,  $m_{soil}$ and $I_{soil}$, too, can be determined by counting the number and distribution of the grains of soil within the bucket.  

In an experimental setting, additional instrumentation will be necessary to obtain these variables in real time. The bucket-soil interaction forces can be estimated based on piston pressure measurements \cite{Jud2019}, or be directly measured by instrumenting the bucket \cite{Johnson2012}. The inertial variables, $m_{soil}$ and $I_{soil}$, can be estimated with use of a vision system. Assuming a constant soil density  $\rho_{soil}$, the soil mass is given by $m_{soil} = \rho_{soil}V_{soil}$ where $V_{soil} = \int dV$ is the volume of all the soil grains in the bucket.  Similarly $I_{soil} = {\rho}_{soil}\int r^2 dV$  where $r$ is the distance from the tip to all the points within the bucket. These soil volume and distribution can be determined from images of the bucket and the surrounding soil.
In recent years, camera systems used to monitor the bucket of excavation machines, mounted on the boom and arm, have been proposed and this technology can be used to estimate those quantities\cite{BoomCamera}. Given a known pose of the bucket and a depth camera attached properly, one can estimate the volume of soil within the bucket as well as evaluate the integral necessary for the inertia. 

The soil surface shape function $s(x)$ is determined by matching a cubic spline to the initial soil surface and using that smooth functional approximation to evaluate the height, $s$ and derivatives $s'$ and $s''$.

\section{Results and Discussion}

This section presents the simulation experiment results obtained from the simulation environment providing all the state and auxiliary variables. 

\subsection{Model Evaluation} 
\begin{figure}[hbt!]
    \centering
    \includegraphics[trim={2.65cm 9.55cm 2.75cm 4.5cm},clip, width=0.99\columnwidth]{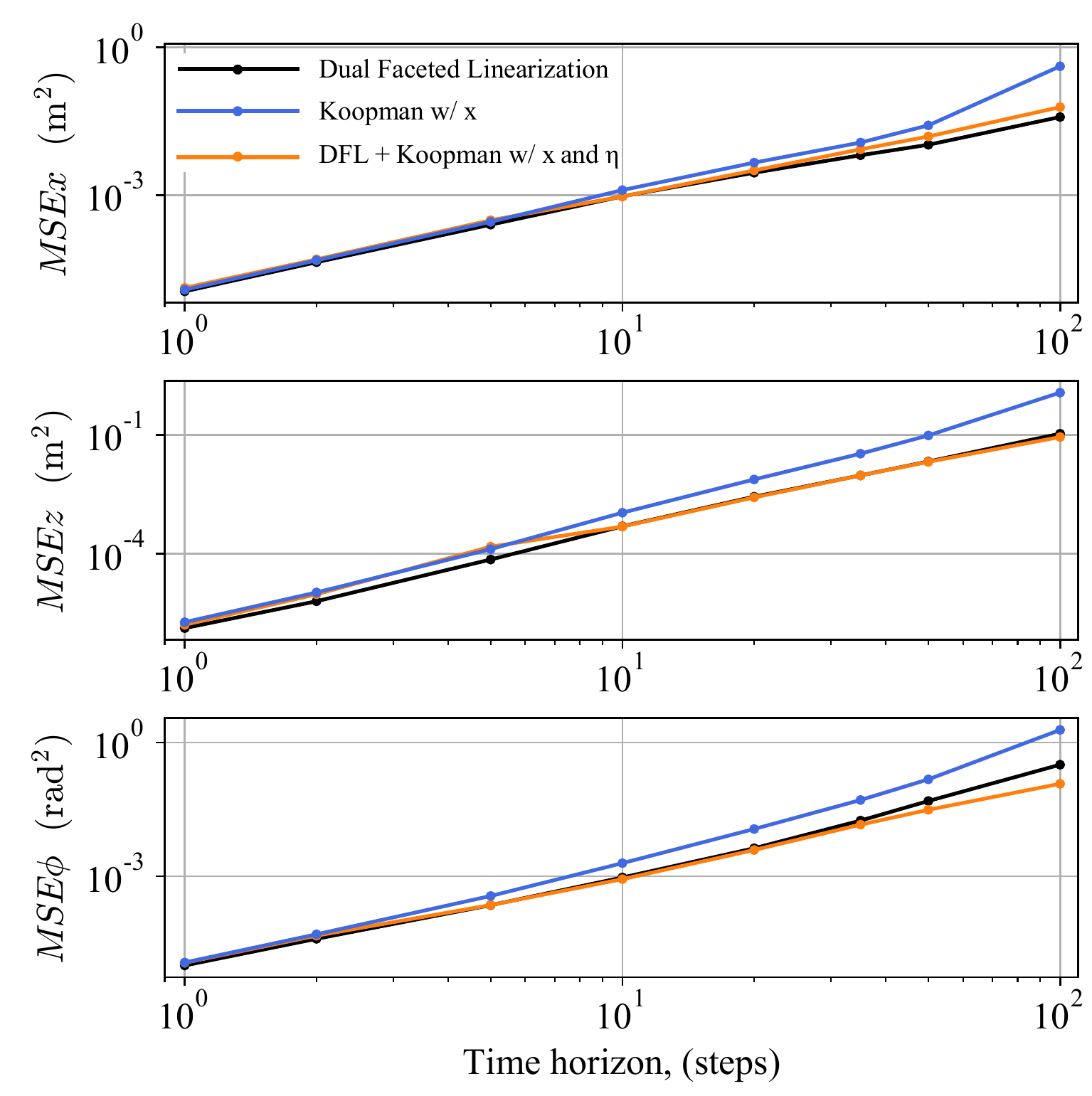}
    \caption{The MSE error for each of the three position variables using DFL and Koopman. The black line is for the DFL model presented in this paper, blue is the conventional Koopman-operator method with polynomial observables of state variables, and orange is the Koopman-operator model utilizing both the independent states and auxiliary variables with polynomial observables. The DFL model (black) is 14th order, the Koopman with 2nd-order polynomial observables of state $\vv{x}$ (blue) is 27th order, and the combined Koopman + DFL with 2nd-order polynomial observables of $\vv{x}$ and $\eta$ (orange) is 119th order.}\label{fig:model_error}
\end{figure}

First, we evaluate the model in terms of prediction accuracy compared to the ground truth agxTerrain simulation values. The Mean Squared Error (MSE) of the bucket position and orientation over an extended time horizon is shown in Fig. \ref{fig:model_error} in logarithmic scale. Each data-point on this plot is based on 30,000 test samples. From the data, 10 different data sets were created and the model was trained for each data set. Furthermore, each model was tested using 10 validation trajectories and forecasting forward from 300 randomly selected initial locations. The plots in Fig. \ref{fig:model_error} show the average of these test results. Prediction error is small in general. However, it increases as the time horizon increases. For MPC application, this lifted linear model can be used as a valid model over a finite time horizon.

The proposed DFL-based lifted linear model was compared to a Koopman-based model, where a set of polynomial functions of state variables were used as observables. We can find that the DFL-based model outperforms the Koopman-based method, although the order of the DFL model (14th order) is much lower than that of the Koopman (27th order).

Note that in DFL \textit{measured} auxiliary variables are used. Even for a considerably smaller number of total observables, the DFL model generally outperforms the Koopman lifted model that utilizes only the state variables (black versus blue data).  
This highlights the importance of judicious choice of measurements with which to lift the dynamics.  Physically meaningful measurements provide critical information for regressing higher accuracy models. 

In Fig. \ref{fig:model_error}, the DFL model was also compared to the Koopman-based model where both state and auxiliary variables were used for constructing observables. The resultant lifted system order of this Koopman model is 119th order. Despite the high order, the Koopman-model does not exhibit significant improvement in prediction accuracy compared to the DFL model.

\begin{figure}[hbt!]
    \centering
    \includegraphics[trim={2.0cm 9.79cm 18.7cm 1.5cm},clip, width=0.9\columnwidth]{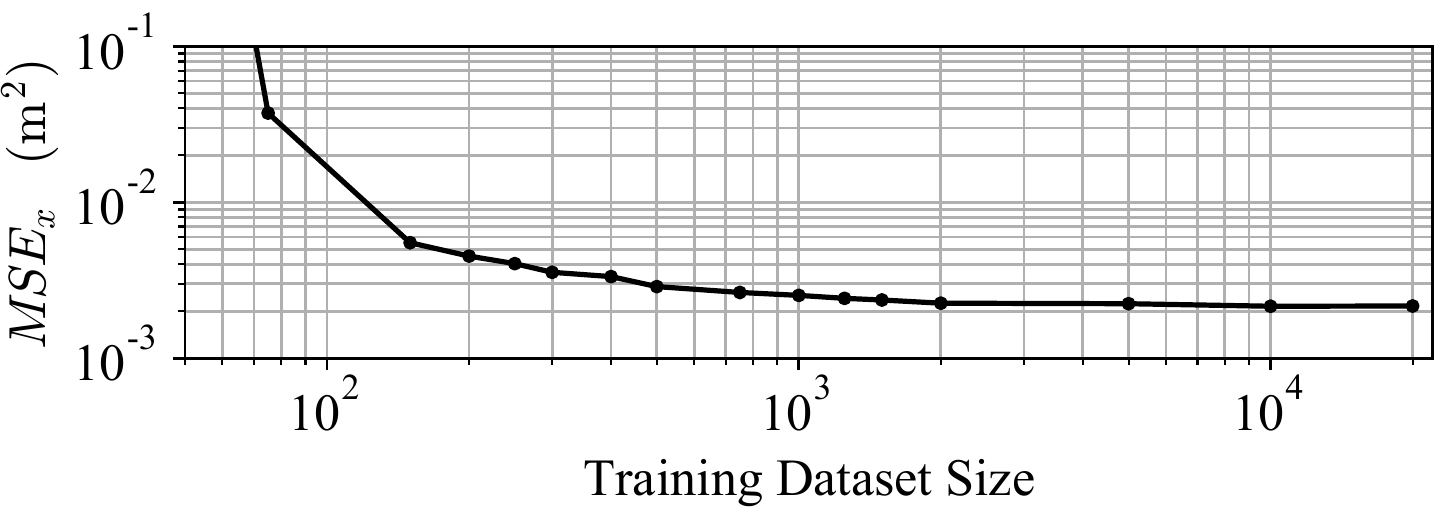}
    \caption{Prediction MSE at a time horizon of 20 steps as a function of training dataset size. For all the variables it is observed that the model prediction improves with increased dataset size but saturates at approximately 2000-3000 sample pairs.}\label{fig:model_error_dataset_size}
\end{figure}

In constructing these models, it was found that a model with fewer variables is more robust to training data-set selection and regression. We found that further increasing the polynomial order of the Koopman models incurred a significant increase of validation test error.  This was observed to be caused by the regressed model often being dynamically unstable. This is an issue which has been observed by prior works \cite{Mamakoukas2020LearningDS, Cibulka2019}. The observables determined through DFL appear to be more robust to obtaining a model from data than higher order models when using a pure L2 regression. The effect of training dataset size shown in Fig. \ref{fig:model_error_dataset_size} revealed that approximately 2,000 sample pairs were sufficient to learn the model to the level where no more data significantly improve performance.

\subsection{Trajectory Control Evaluation}
\begin{figure}[hbt!]
    \centering
    \includegraphics[trim={1.5cm 13.5cm 3.9cm 4.85cm},clip, width=0.99\columnwidth]{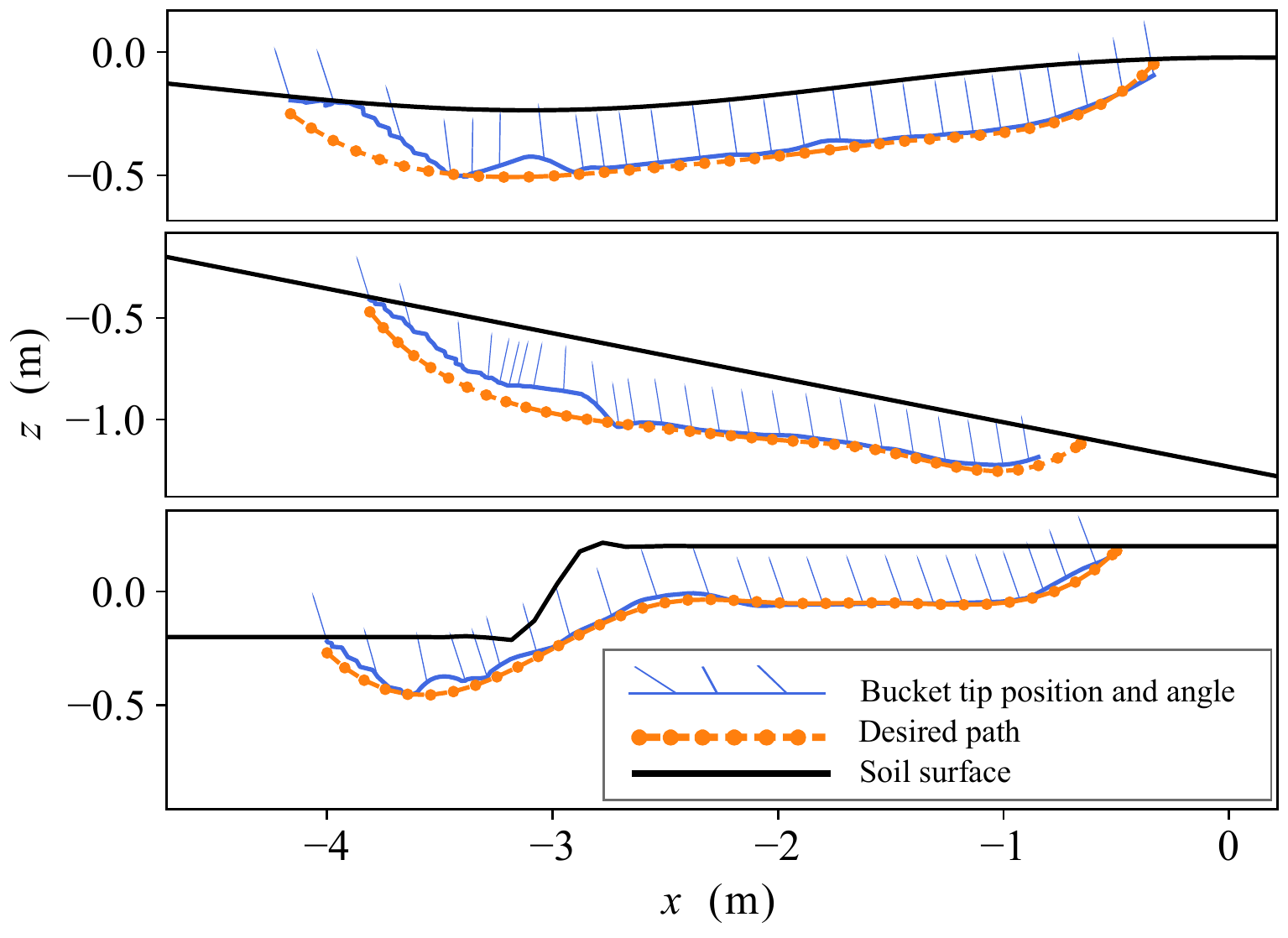}
    \caption{Several sample results using the MPCC-DFL controller on soil with various shapes.}\label{fig:trajectories}
\end{figure}

\begin{figure}[hbt!]
    \centering
    \includegraphics[trim={.5cm 22.5cm 1.85cm 1.55cm},clip, width=0.99\columnwidth]{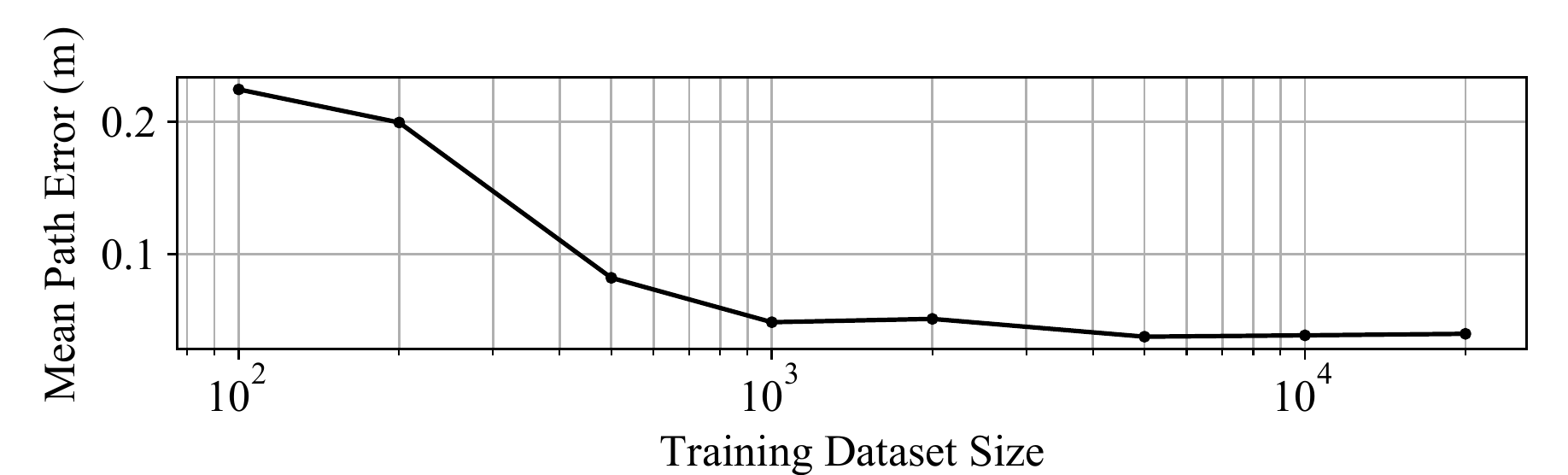}
    \caption{The mean path error utilizing the proposed controller, with low $q_\theta$ such that a close following of the path is sought is plotted for varying dataset size used to train the DFL model. The path error is defined as the minimum distance between a $(x,z)$ location and the path.}\label{fig:model_error_dataset_size_control}
\end{figure}

We also evaluated the performance of the proposed model for contouring control of the bucket system along desired paths. The MPCC control was implemented for the DFL-based lifted linear model and tested in the agxTerrain simulation environment. Fig. \ref{fig:trajectories} shows the simulation experiment results of the tip trajectories for several soil profiles. The bucket successfully follows the path, despite the initial transient error. Once the bucket tip arrived near the desired path, the MPCC controller allowed the bucket to track the desired path although the soil profile varied. 
The path tracking accuracy was improved as the model was trained with a larger training dataset, as shown in Fig. \ref{fig:model_error_dataset_size_control}. With 
a high contouring error penalty, the mean path error decreased with increasing training dataset size. The performance appears to saturate at approximately the same dataset size as in the case of the model prediction performance. 

\begin{figure}[bt!]
    \centering
    \includegraphics[trim={2.cm 19.25cm 2.8cm 4cm},clip, width=0.99\columnwidth]{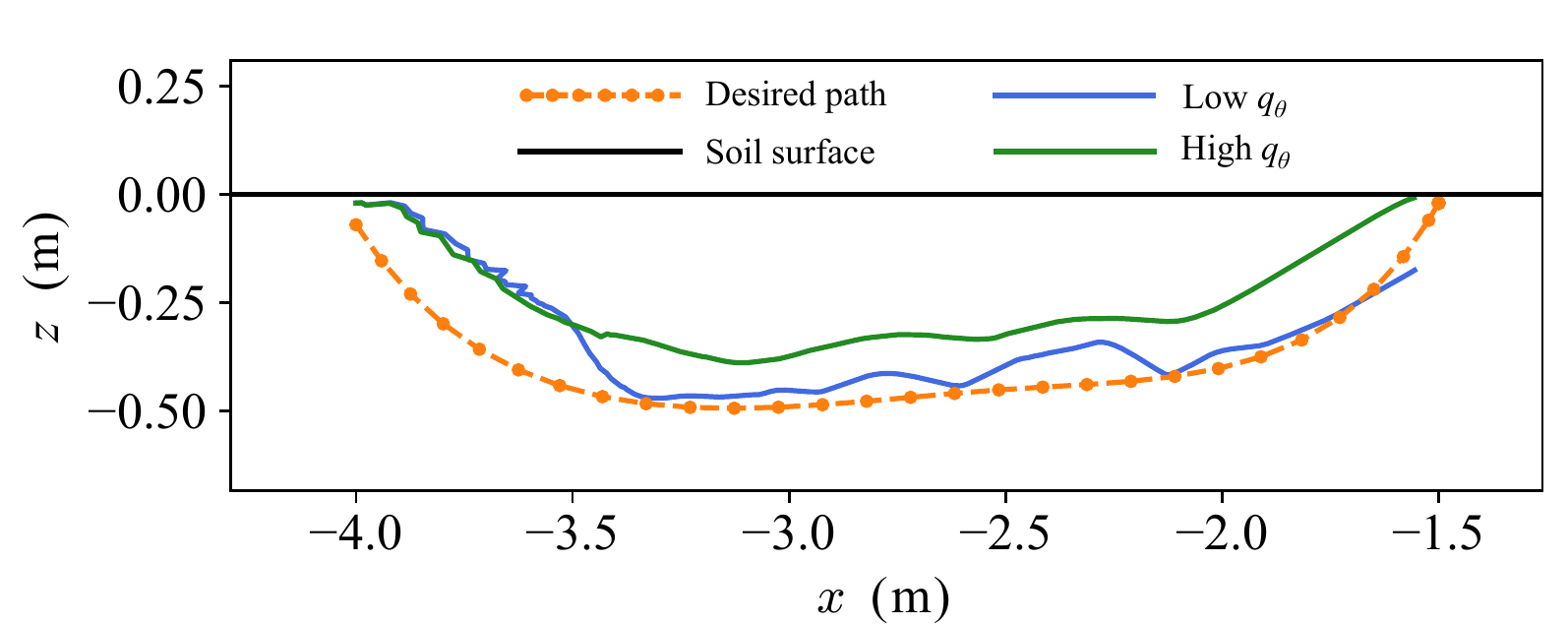}
    \caption{Changing the path progression weight $q_\theta$ changes the balance of progressing rapidly versus accurately through the soil. For lower values of $q_\theta$ the path is followed more precisely but at a slower speed. For this trial the time to complete the excavation cycle was 4.2s for $q_\theta=4$ (green) and 7.0s for $q_\theta=1$ (blue).}\label{fig:trajectories_q}
\end{figure}

Changing the relative magnitude of the weight matrices in the MPCC controller allows us to make the trade-off between speed progression along the path and precision of the digging task. As illustrated in Fig. \ref{fig:trajectories_q},  modulating $q_\theta$ alters the resulting trajectory and time to completion, for identical soil conditions and desired paths. 

\begin{figure}[bt!]
    \centering
    \includegraphics[trim={2cm 13.0cm 2cm 5.1cm},clip, width=0.95\columnwidth]{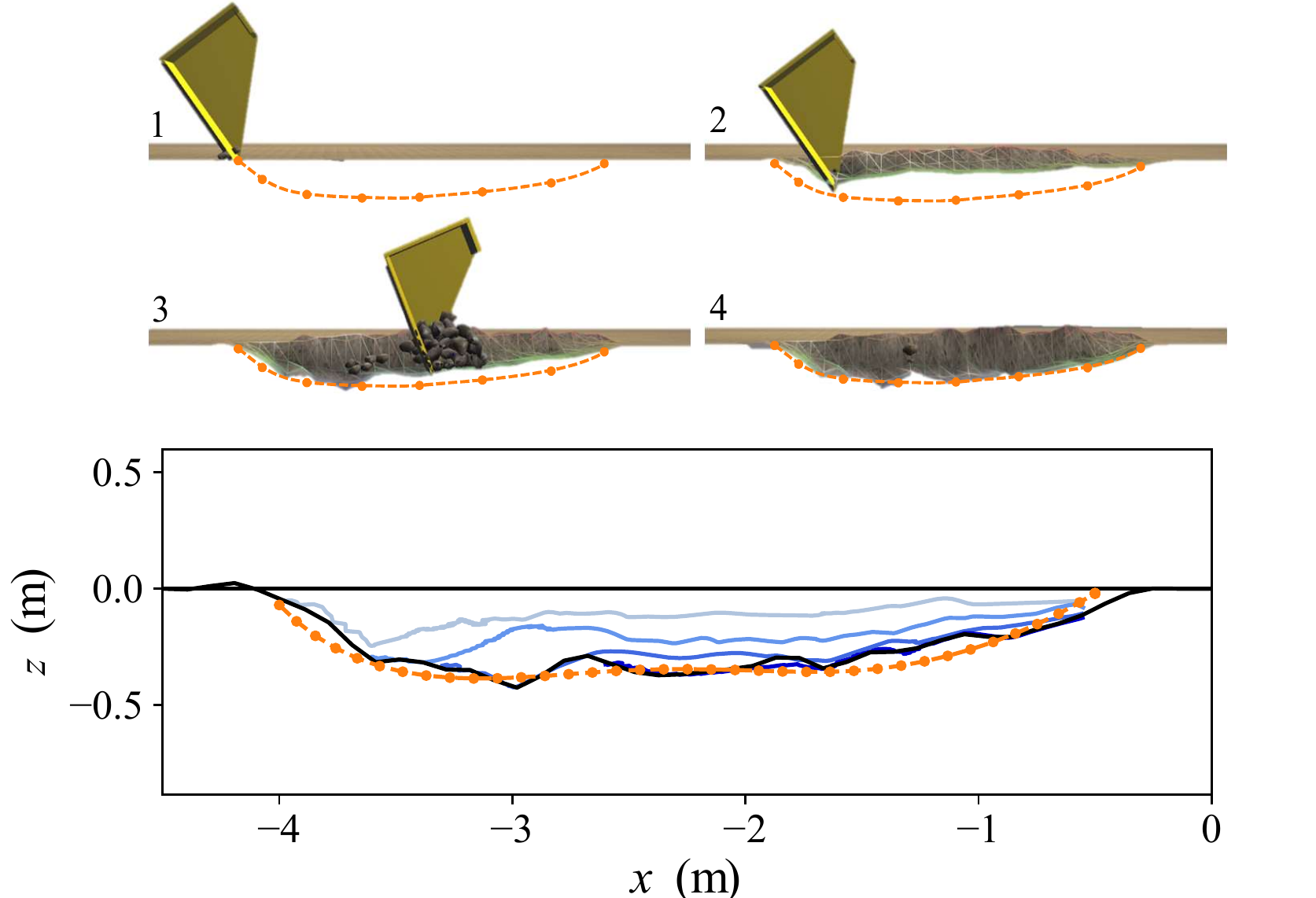}
    \caption{Excavation of a desired  soil shape is achieved through consecutive digging cycles utilizing the same path and control parameters. The black lines indicate the initial and final soil shape, the orange the desired shape, and the progressively darker blue lines the successive digging trajectories.}\label{fig:trajectories_multiscoop}
\end{figure}

It is interesting to note that the proposed MPCC controller can also be utilized to incrementally find a multi-path excavation trajectory for digging a deep trench. See Fig. \ref{fig:trajectories_multiscoop}. As the profile becomes deep, it is impossible to excavate it in a single scooping cycle due to input or soil force constraints. Instead, the optimization subject to these constraints leads to incremental multi-cycle excavation. Using the same path and same control parameters the soil profile is excavated through consecutive cycles. 

\section{Conclusion and Further Work}
In this paper we have demonstrated the feasibility of using lifting linearization to model the dynamic interactions between a bucket and the soil it is excavating. The learned DFL model was compact, yet capable of capturing a significant portion of the nonlinear dynamics. Using this model we found that we were able to control a bucket to follow desired paths, while respecting state and input constraints, using a model predictive contouring controller with quadratic cost and linear constraints. 

There are a few important issues to be addressed further. First, the proposed method must be integrated with a complete excavator controller. This paper focused on the bucket-soil dynamics and omitted the dynamics of the hydraulically-powered excavator dynamics. MPC control of hydraulic excavators has been addressed in multiple prior works \cite{Bender2017,Tomatsu2016}. It is necessary to integrate the bucket-soil dynamic control with those prior works focusing on the excavator control with simple soil dynamics. 

Another critical aspect which must be addressed is dealing with varying or changing soil types. Currently we rely on a static dataset which is trained on a single soil type. By leveraging techniques in active learning for lifted linearization \cite{Abraham2019} or selectively sampling from a large pre-existing dataset we may be able to expand the applicability to varying soil properties encountered by the excavator.

%
%
\FloatBarrier
\bibliographystyle{IEEEtran}
\bibliography{IEEEabrv,biblio}
\end{document}